\pdfoutput=1

\documentclass[11pt]{article}

\usepackage{acl}

\PassOptionsToPackage{dvipsnames,table}{xcolor}
\usepackage{times}
\usepackage{latexsym}

\usepackage{colortbl}
\usepackage{booktabs}
\usepackage{multirow}
\usepackage{rotating}
\usepackage{makecell}
\usepackage{lipsum} 
\usepackage{graphicx}
\usepackage{float}
\usepackage{caption,subcaption}
\usepackage[export]{adjustbox}
\usepackage[tableposition=top]{caption}
\usepackage{arydshln}

\usepackage[T1]{fontenc}

\usepackage[utf8]{inputenc}

\usepackage{microtype}

\title{Is More Data Better? Re-thinking the Importance of Efficiency in Abusive Language Detection with Transformers-Based Active Learning}

\author{Hannah Rose Kirk$^{1, 2, 3 \ddagger}$, Bertie Vidgen$^{1, 3}$, Scott A. Hale$^{1, 2, 4}$\\
  { $^1$University of Oxford, $^2$The Alan Turing Institute, $^3$Rewire, $^4$Meedan} \\
  { $^\ddagger$hannah.kirk@oii.ox.ac.uk}\\
}
  
\usepackage{cleveref}
\crefname{section}{Sec.}{Sec.}
\crefname{Section}{Sec.}{Sec.}
\crefname{table}{Tab.}{Tab.}
\crefname{appendix_table}{Tab.}{Tab.}
\crefname{Table}{Tab.}{Tab.}
\crefname{Figure}{Fig.}{Fig.}
\crefname{figure}{Fig.}{Fig.}
\crefname{appendix}{Appendix}{Appendix}
\crefname{chapter}{Chapter}{Chapter}

\usepackage{fancyhdr}

\fancypagestyle{firstpage}
{
    \fancyhead[L]{}    
    \fancyhead[R]{Paper accepted for publication at the \textit{3rd Workshop on Threat, Aggression and Cyberbullying (COLING 2022)}}
}

\begin{document}
\thispagestyle{firstpage}
\maketitle
\begin{abstract}
Annotating abusive language is expensive, logistically complex and creates a risk of psychological harm. However, most machine learning research has prioritized maximizing \textit{effectiveness} (i.e., F1 or accuracy score) rather than data \textit{efficiency} (i.e., minimizing the amount of data that is annotated).
In this paper, we use simulated experiments over two datasets at varying percentages of abuse to demonstrate that transformers-based active learning is a promising approach to substantially raise efficiency whilst still maintaining high effectiveness, especially when abusive content is a smaller percentage of the dataset. This approach requires a fraction of labeled data to reach performance equivalent to training over the full dataset.
\end{abstract}

\section{Introduction}
Online abuse, such as hate and harassment, can inflict psychological harm on victims \cite{Gelber2016harms}, disrupt communities \cite{Mohan2017communities} and even lead to physical attacks \cite{Williams2019offline}. Machine learning solutions can be used to automatically detect abusive content at scale, helping to tackle this growing problem \cite{Gillespie2020a}.
An \textit{effective} model is one which makes few misclassifications, minimizing the risk of harm from false positives and negatives: false negatives mean that users are not fully protected from abuse while false positives constrain free expression.
Most models to automatically detect abuse are trained to maximize effectiveness via ``passive'' supervised learning over large labeled datasets. However, although collecting large amounts of social media data is relatively cheap and easy, annotating data is expensive, logistically complicated and creates a risk of inflicting psychological harm on annotators through vicarious trauma \cite{robertsScreenContent2019, Steiger2021}. 
Thus, an \textit{efficient} model, which achieves a given level of performance with few labeled examples, is highly desirable for abusive content detection.

\begin{figure}[!t]
    \centering
    \includegraphics[width = \columnwidth]{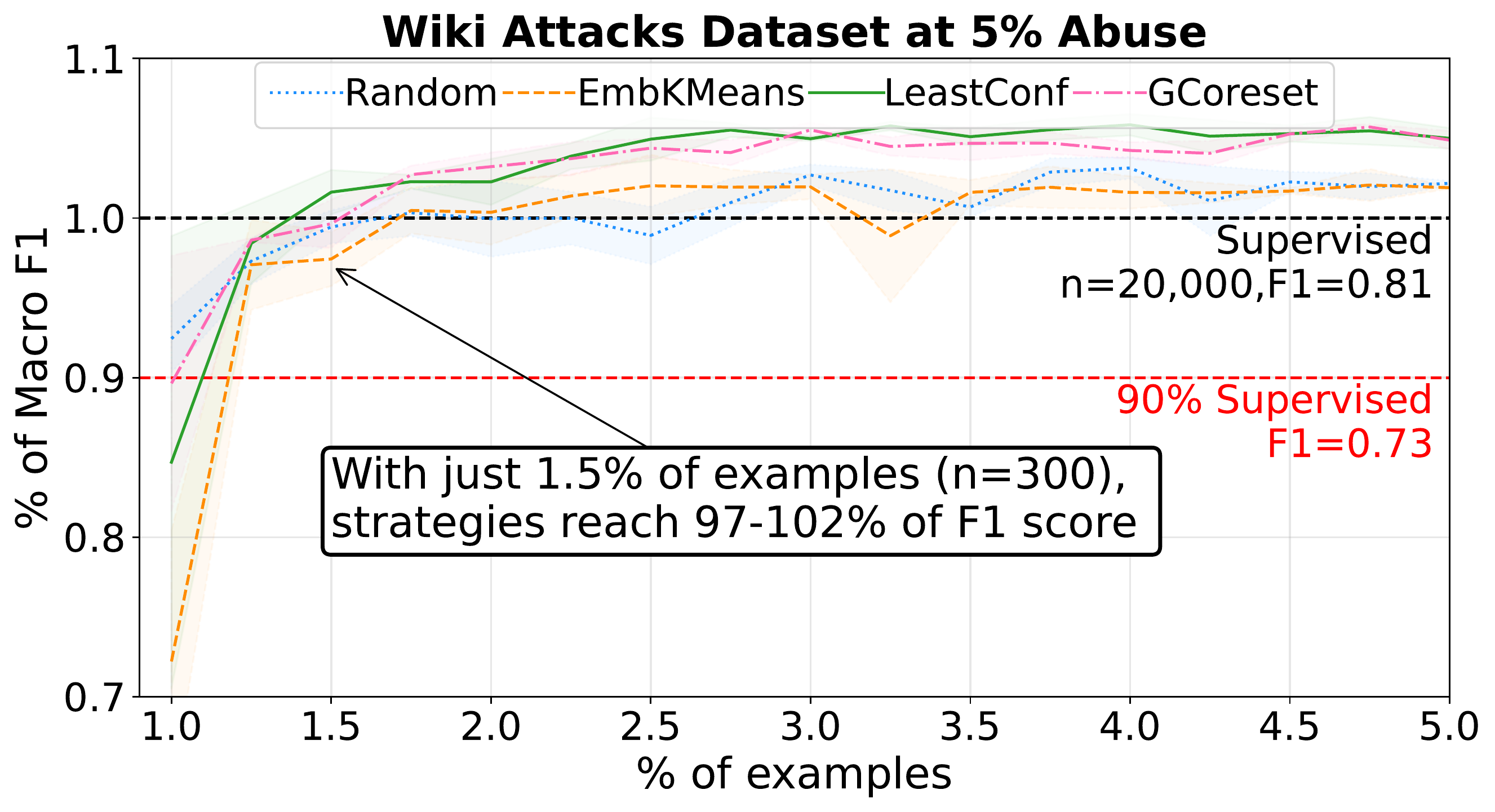}
    \caption{Transformers-based active learning beats fully-supervised baseline with $1.5\%$ of the 20,000 examples.}
    \label{fig:splash}
\end{figure}

Our central objective is to demonstrate how to maximize efficiency and effectiveness when training abuse detection systems, and in this paper, we focus on active learning (AL).
AL is an iterative human-in-the-loop approach that selects entries for annotation only if they are `informative' \cite{Lewis1994, Settles2009}. While AL has shown promise for abusive language dataset creation \cite{Charitidis2020, Mollas2020, Rahman2021, Bashar2021, Abidin2021}, there are several open questions about the most appropriate configuration and use. In particular, only one paper uses transformers-based AL for abusive language detection \cite{EinDor2020} to our knowledge, although the benefits of AL for other classification tasks is clear \cite{Schroder2021AL, EinDor2020, Yuan2020}.
Pre-trained transformer models have been widely adopted for abuse detection, but while they can be fine-tuned on relatively few examples for specific tasks \cite{Devlin2018bert, Qiu2020survey}, they are still commonly used with large datasets \cite[e.g.][]{mozafari2019bert, mutanga2020hate, isaksenUsingTransferbased2020, koufakouHurtBERTIncorporating2020}.
Our first subquestion asks,
\textbf{RQ1.1:} \textit{What effect do model pre-training and architecture have on efficiency and effectiveness?}
To answer RQ1.1, we evaluate transformers- and traditional-based AL in a simulated setup using two already-labeled abusive language datasets. 

One challenge in abusive language detection is class imbalance as, although extremely harmful, abuse comprises a small portion of online content \cite{Vidgen2019howmuch}. Prior AL work primarily uses datasets at their given class imbalances and thus has not disentangled how class imbalance versus linguistic features affect the design choices needed for efficient AL. This is a problem given that most abusive language datasets do not reflect the imbalance actually observed in the wild.
Our second subquestion addresses this issue, \textbf{RQ1.2:} \textit{What effect does class imbalance have on efficiency and effectiveness?}
To answer RQ1.2, we artificially-rebalance the datasets at different percentages of abuse. 

In addressing these questions, we find that more data is not always better and can actually be worse, showing that effectiveness and efficiency are not always in tension with one another.
With extensive pre-training and greater model complexity, a transformers-based AL approach achieves high performance with only a few hundred examples. 
Crucially, we show that the value of transformers-based AL (relative to random sampling) is larger for more imbalanced data (i.e., data that more closely reflects the real-world). 
For 5\% abuse, the performance of a transformers-based AL strategy over 3\% of a 20k dataset can even surpass the F1 of a model passively trained over the full dataset by 5 percentage points (\cref{fig:splash}). In \S\ref{sec:discussion} we describe caveats of our findings and implications for future research in abusive language detection.\footnote{Code at \href{https://github.com/HannahKirk/ActiveTransformers-for-AbusiveLanguage}{ActiveTransformers-for-AbusiveLanguage}.}

\section{Methods}
\subsection{Active Learning Set-Up}
AL typically consists of four components: 1) a classification model, 2) pools of unlabeled data $\mathcal{U}$ and labeled data $\mathcal{L}$, 3) a query strategy for identifying data to be labeled, and 4) an `oracle' (e.g., human annotators) to label the data.
First, seed examples are taken from $\mathcal{U}$ and sent to the oracle(s) for labeling. These examples initialize the classification model. Second, batches of examples are iteratively sampled from the remaining unlabeled pool, using a query strategy to estimate their `informativeness' to the initialized classification model.\footnote{Note that \textit{batch-mode active learning} is a common application in both research and industry, given its more practical application to annotation workflows and model retraining times \cite[p.~35]{Settles2009}.} Each queried batch is labeled and added to $\mathcal{L}$. Finally, the classifier is re-trained over $\mathcal{L}$.\footnote{We train from scratch to avoid overfitting to previous iterations \cite{EinDor2020, Hu2018}.} 

\begin{table}
\centering
\footnotesize
\caption{Summary of source datasets (in gray) and their artificially-rebalanced versions.}
\label{tab:summary_datasets}
\setlength{\tabcolsep}{2pt}
\resizebox{\columnwidth}{!}{
\begin{tabular}{cc cc cc}
\toprule
& & \multicolumn{2}{c}{\textbf{Train$^\dagger$}} & \multicolumn{2}{c}{\textbf{Test$^*$}}    \\ 
\cmidrule(lr){3-4}\cmidrule(lr){5-6}
                                   \textbf{Dataset}       & \textbf{Imbalance} & \textbf{abuse} & \textbf{non-abuse} & \textbf{abuse} & \textbf{non-abuse}  \\ 
\midrule
\rowcolor[rgb]{0.8,0.8,0.8} \textbf{wiki}   & 12\%               & 10,834         & 81,852             & 2,756           & 20,422              \\
\textbf{wiki50}                             & 50\%               & 10,000         & 10,000             & 2,500           & 2,500                \\
\textbf{wiki10}                             & 10\%               & 2,000           & 18,000             & 500            & 4,500                \\
\textbf{wiki5}                              & 5\%                & 1,000           & 19,000             & 250            & 4,750                \\ 
\midrule
\rowcolor[rgb]{0.8,0.8,0.8} \textbf{tweets} & 32\%               & 28,955         & 61,041             & 3,160           & 6,840                \\
\textbf{tweets50}                           & 50\%               & 10,000         & 10,000             & 2,500           & 2,500                \\
\textbf{tweets10}                           & 10\%               & 2,000           & 18,000             & 500            & 4,500                \\
\textbf{tweets5}                            & 5\%                & 1,000           & 19,000             & 250            & 4,750                \\
\bottomrule
\multicolumn{1}{l}{\textit{Notes:}} & \multicolumn{5}{r}{$^\dagger$ Train is used as the unlabeled pool ($n=20{,}000$)} \\
\multicolumn{1}{l}{\textit{}} & \multicolumn{5}{r}{$^*$ Test is used for held-out evaluation ($n=5{,}000$)}
\end{tabular}
}
\end{table}

\subsection{Dataset Selection and Processing}
AL is path-dependent---i.e., later decisions are dependent upon earlier ones; so, experimenting in real-world settings is prohibitively costly and risky to annotator well-being. To reproduce the process without labeling new data, we use existing labeled datasets but withhold the labels until the model requests their annotation.
We examined a list of publicly available, annotated datasets for abusive language detection\footnote{\url{https://hatespeechdata.com}} and found two that were sufficiently large and contained enough abusive instances to facilitate our experimental approach. The \textbf{wiki} dataset \cite{Wulczyn2017} contains comments from Wikipedia editors, labeled for whether they contain personal attacks. A test set is pre-defined; we take our test instances from this set. The \textbf{tweets} dataset \cite{Founta2018} contains tweets which have been assigned to one of four classes. We binarize by combining the abusive and hate speech classes (=1) and the normal and spam classes (=0) to allow for cross-dataset comparison \cite{wiegandDetectionAbusive2019, EinDor2020}. A test set is not pre-defined; so, we set aside 10\% of the data for testing that is never used for training.

To disentangle the merits of AL across class imbalances, we construct three new datasets for both \textbf{wiki} and \textbf{tweets} that have different class distributions: 50\% abuse, 10\% abuse and 5\% abuse. This creates 6 datasets in total (see \cref{tab:summary_datasets}).
To control dataset size and ensure we have sufficient positive instances for all imbalances, we assume that each unlabeled pool has $20{,}000$ examples.\footnote{The \textbf{wiki} dataset has 10,834 abusive entries; so, at 50\% abuse, the upper limit on a rebalanced pool is $21{,}668$.} We experiment with multiple AL strategies to select $2{,}000$ examples for annotation as early experiments showed further iterations did not affect performance.\footnote{AL experiments are implemented in the Python \texttt{small-text} library \cite{Schroder2021smalltext}} 

\subsection{Experimental Setup}
We use 2 model architectures, 2 query strategies and 6 artificially-rebalanced datasets, giving 24 experiments each of which we repeat with 3 random seeds. Each experiment uses the same sized unlabeled pool, training budget and test set (see \cref{tab:summary_datasets}).
In figures, we present the mean run (line) and standard deviation (shaded). For transformers-based AL, we use distil-roBERTa (\textbf{dBERT}), which performs competitively to larger transformer models \cite{sanh2019distilbert}, also in an AL setting \cite{Schroder2021AL}. For traditional AL without pre-training, we use a linear support vector machine (\textbf{SVM}) as a simple, fast and lightweight baseline.\footnote{\cref{sec:app_model_training} presents details of model training.} For active data acquisition, we try three AL strategies; LeastConfidence, which selects items close to the decision boundary \cite{Lewis1994}, is presented in the paper while the other strategies are in the Appendix.\footnote{We also test GreedyCoreSet \cite{Sener2017} and EmbeddingKMeans \cite{Yuan2020}, but LeastConfidence outperformed them.} For comparison, we randomly sample items from the unlabeled pool at each iteration. Alongside model and query strategy, AL requires an initial seed size, seed acquisition strategy and batch size. We experimentally determined the best values for these parameters: an initial seed of 20 examples selected via a keyword-heuristic \cite{EinDor2020} and batches of 50 examples.\footnote{We present pilot experiments in \cref{sec:app_keywords} and \ref{sec:app_additional}.}

\subsection{Evaluation}
As a baseline, we use the passive macro-F1 score over the full dataset of $20{,}000$ entries ($\mathbf{F1_{20k}}$). For each AL strategy, we measure efficiency on the held-out test set as the number of examples needed to surpass 90\% of $\mathbf{F1_{20k}}$, which we call $\mathbf{N_{90}}$.\footnote{To fairly compare models, we calculate $\mathbf{N_{90}}$ relative to \textit{best} $\mathbf{F1_{20k}}$ (achieved by dBERT in all cases).} For effectiveness, we use the maximum F1 score achieved by each AL strategy, which we call $\mathbf{F1_{AL}}$.

\section{Results}

\begin{table}
\centering
\caption{Efficiency and effectiveness of each classifier (transformers vs SVM) with LeastConfidence sampling.}
\label{tab:results_summary_table}
\footnotesize
\begin{tabular}{cc|ccc} 
\toprule
\textbf{Dataset}                    & \textbf{Classifier} & \textbf{$\mathbf{F1_{20k}}^\dagger$} & \textbf{$\mathbf{F1_{AL}}$} & \textbf{$\mathbf{N_{90}}$}  \\ 
\hline
\multirow{2}{*}{\textbf{wiki50}}    & dBERT               & \textbf{0.920}                       & \textbf{0.920}              & \textbf{170}                \\
                                    & SVM                 & 0.875                                & 0.836                       & 1570                        \\ 
\hline
\multirow{2}{*}{\textbf{wiki10}}    & dBERT               & \textbf{0.859}                       & \textbf{0.866}              & \textbf{170}                \\
                                    & SVM                 & 0.809                                & 0.810                       & 320                         \\ 
\hline
\multirow{2}{*}{\textbf{wiki5}}     & dBERT               & \textbf{0.807}                       & \textbf{0.855}              & 220                         \\
                                    & SVM                 & 0.785                                & 0.780                       & \textbf{170}                \\ 
\hline
\multirow{2}{*}{\textbf{tweets50}}  & dBERT               & \textbf{0.939}                       & \textbf{0.938}              & \textbf{170}                \\
                                    & SVM                 & 0.931                                & 0.926                       & 220                         \\ 
\hline
\multirow{2}{*}{\textbf{tweets10}}  & dBERT               & \textbf{0.904}                       & \textbf{0.902}              & 220                         \\
                                    & SVM                 & 0.893                                & 0.901                       & \textbf{170}                \\ 
\hline
\multirow{2}{*}{\textbf{tweets5}}   & dBERT               & \textbf{0.844}                       & \textbf{0.856}              & 300                         \\
                                    & SVM                 & 0.825                                & 0.830                       & \textbf{170}                \\ 
\bottomrule
\multicolumn{5}{l}{\scriptsize \textit{Notes:}$^\dagger$global metric from passive training over full, re-balanced dataset}
\end{tabular}
\end{table}
\label{sec:results}
\paragraph{Efficiency \& Effectiveness} For each dataset, we find active strategies that need just 170 examples (0.8\% of the full dataset) to reach 90\% of passive supervised learning performance (see \cref{tab:results_summary_table}). When training over the full dataset, dBERT always outperforms SVM, models have worse performance on more imbalanced datasets, and \textbf{wiki} is harder to predict than \textbf{tweets} (\cref{tab:results_summary_table}). In all cases, LeastConfidence outperforms the random baseline, and the gain is larger for lower percentages of abuse: for \textbf{wiki10} and \textbf{wiki5}, $\mathbf{N_{90}}$ is lower by 150 and 100 examples, respectively. AL can even outperform passive supervised learning over the full dataset, showing there is no efficiency--effectiveness trade-off. For the majority of datasets, dBERT with LeastConfidence over $2{,}000$ examples matches or surpasses the F1 score of a model trained passively over the whole dataset ($\mathbf{F1_{AL}} \ge \mathbf{F1_{20k}}$ in \cref{tab:results_summary_table}). For \textbf{wiki5}, it is 5 percentage points (pp) higher (\cref{fig:splash}).

\begin{figure*}[!t]
\minipage[t]{0.47\textwidth}
  \includegraphics[width=\linewidth]{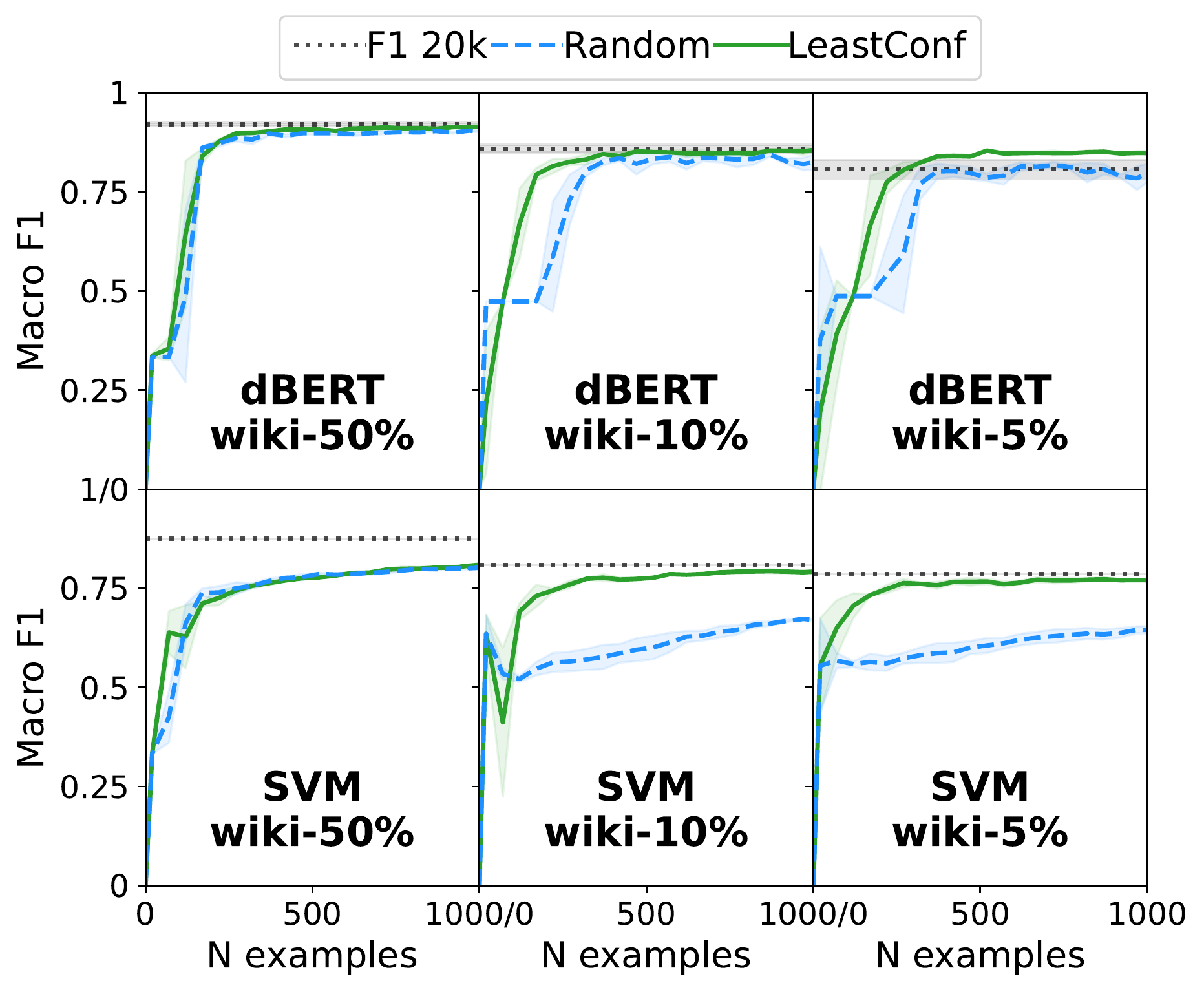}
  \caption{\centering The contribution of pre-training vs active data acquistion.}\label{fig:model_query}
\endminipage\hfill
\minipage[t]{0.47\textwidth}
  \includegraphics[width=\linewidth] {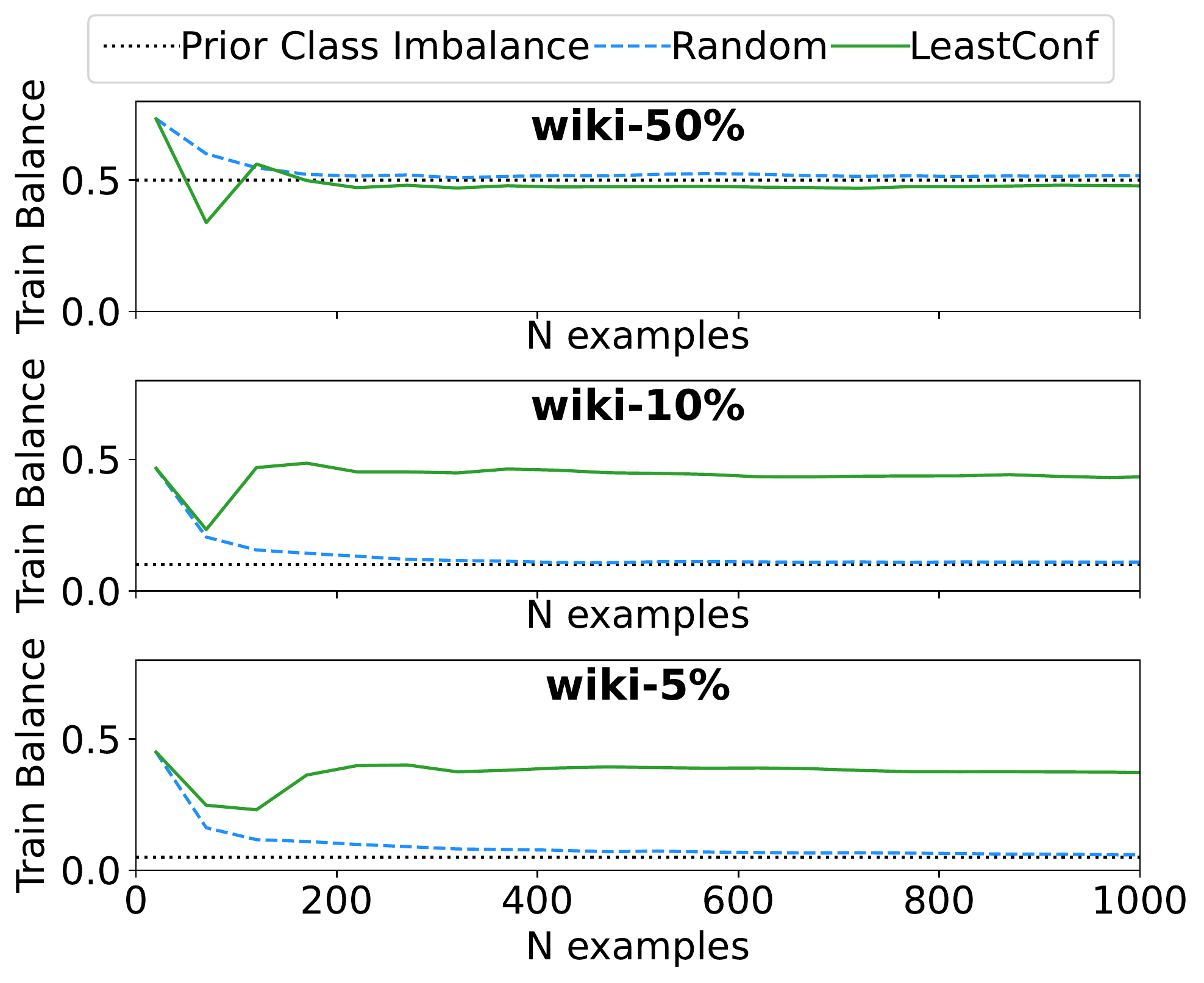}
  \caption{\centering Label imbalance during training (dBERT).}\label{fig:train_imbalance}
\endminipage
\hfill
\end{figure*}

\paragraph{The Effect of Pre-Training} We find AL  has a bigger impact for SVM than dBERT, shown by the larger gap to the random baselines (\cref{fig:model_query}). With its extensive pre-training, dBERT achieves high performance with few examples, even if randomly selected. Nonetheless, an AL component still enhances dBERT performance above the random baseline especially with imbalanced data \cite[as found by][]{Schroder2021AL, EinDor2020}, requiring 150 and 100 fewer examples for $\mathbf{N_{90}}$, and raising F1 score by 2pp and 4pp, for \textbf{wiki5} and \textbf{wiki10} respectively.

\paragraph{Train Distribution} To assess why AL is more impactful with imbalanced data, we evaluate the distribution of the labeled pool at each iteration (\cref{fig:train_imbalance}). The random baseline tends to the original distribution as expected but the LeastConfidence strategy actively selects abusive examples from the pool and tends toward a balanced distribution.

\begin{figure}[bt]
    \centering
    \includegraphics[width = \columnwidth]{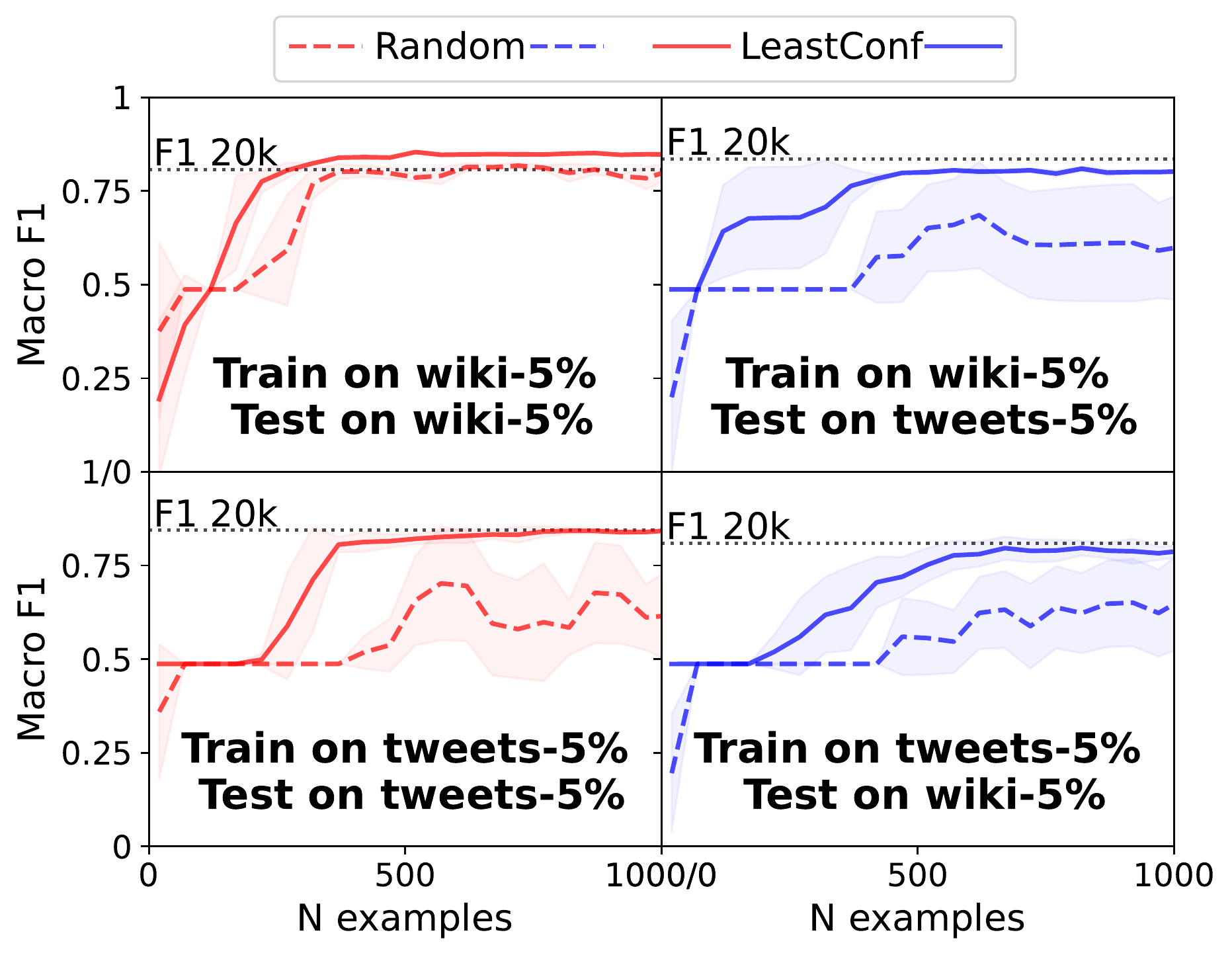}
    \caption{\centering Cross-dataset generalization (dBERT).}
\label{fig:cross}
\end{figure}

\paragraph{Out-of-domain Testing}
The high performance of models trained on few examples raises a risk that they are overfitting and may not generalize. We take the models trained on each of the three class imbalances for \textbf{wiki} and test them on their equivalent \textbf{tweets} dataset, and vice versa. As with in-domain results, models trained on \textbf{wiki} and applied to \textbf{tweets} reach $\mathbf{F1_{20k}}$ within few iterations. The gap between LeastConfidence and the random baseline is larger for out-of-domain evaluation versus in-domain (\cref{fig:cross}). A similar pattern occurs for other imbalances (see \cref{sec:app_generalize}). This suggests that our results for these two datasets are not overfitting.

\section{Discussion}
\label{sec:discussion}
In response to our central research objective, we find strategies which are both effective and efficient, requiring far fewer examples to reach performance equivalent to passive training over the full dataset. These results suggest that passive approaches may be needlessly expensive and place annotators at unnecessary risk of harm. For RQ1.1, we find that coupling pre-trained transformers with AL is a successful approach which leverages the benefits of careful training data selection with the previously demonstrated strong capabilities of pre-trained language models for few-shot learning \cite{brownLanguageModels2020, gaoMakingPretrained2021, schickItNot2021}. However, the compute required to fine-tune a new transformer model in each iteration means AL may have a large environmental footprint \cite{Bender2021}. In some instances, SVMs with AL produce competitive results and have smaller environmental costs. For RQ1.2, we find transformers-based AL is particularly valuable under more extreme class imbalance because it iteratively balances the distribution. Our findings are subject to some limitations, which present avenues for future work.

\paragraph{How does data sampling, class labels and linguistic diversity affect performance?} We evaluate against two datasets with pre-existing labels, which we simplify into a binary task. This binarization was required to allow comparison across datasets. The \textbf{wiki} dataset samples banned comments and \textbf{tweets} samples with keywords and sentiment analysis. While these datasets were the only publicly-available datasets large enough for this work, \citet{wiegandDetectionAbusive2019} shows that they lack diversity, contain numerous biases, and cover abuse which is mostly explicit. This may make it easier for models to learn the task and generalize in fewer examples. Future work should evaluate the success and generalizability of AL for fine-grained labels and implicit abuse.

\paragraph{How does the number of model parameters affect performance?} For computational feasibility and environmental concerns, we use distil-BERT but future work could assess if larger transformers models set higher baselines from passive training over the full dataset.

\paragraph{Are certain AL strategies well-suited to abusive language detection?} We evaluate three commonly-used AL strategies, finding that LeastConfidence performs best, but none are tailored explicitly to abusive language. Constrastive Active Learning \cite{margatina2021active} may be particularly useful: by finding linguistically similar entries on either side of the decision boundary, it may prevent overfitting to certain slurs, profanities or identities.

\paragraph{Do the experimental findings generalize to real-world settings?} Our motivation for maximizing efficiency is to reduce financial costs and risk of harm to annotators, which we operationalize in terms of the number of labeled examples they view. In practice, costs are variable because entries which are more `uncertain' to the model may also be more time-consuming, challenging or harmful for humans to label \cite{haertelAnalyticEmpirical2015}. In a real-world setting, the work of the human annotators must be scaled up and down in response to labeling demands, which may incur additional costs. Crowd-sourced annotators can provide labels on demand when a new batch of entries is launched. With an expert annotation team, there may be a cost of paying annotators during re-training. Furthermore, it is important to note that the scope and scale of realized harm depends on both the total number of annotators as well as their identity, positionality and working conditions. While our approach simulates the labeling process with one groundtruth label, we make no assumptions on how this groundtruth is obtained---either via a single annotator or with some aggregation function over multiple annotator votes---so, our method is applicable to any number or constitution of annotators. We only make the light assumption that less exposure to harm is a good thing---whether that is many people being exposed a little less or few people being exposed a lot less. Future work is needed beyond our simulated set-up to calculate a more realistic cost-benefit ratio of AL, both in terms of financial and psychological costs.

We are exploring these questions in future work but simultaneously encourage the community to consider the need for efficiency in abusive language detection because of the costs, complexities and risk of harm to annotator well-being from inefficient data labeling.

\section*{Acknowledgments}
This work was supported in part by JADE: Joint Academic Data science Endeavour - 2 under the EPSRC Grant EP/T022205/1, and by the The Alan Turing Institute under EPSRC grant EP/N510129/1.
Hannah Rose Kirk was supported by the Economic and Social Research Council grant ES/P000649/1. We thank Paul Röttger and our anonymous reviewers for their comments.

\bibliography{references}
\bibliographystyle{acl_natbib}

\appendix
\cleardoublepage
\section{Details of Dataset Processing and Model Training}
\label{sec:app_model_training}
We use two English-language datasets which were curated for the task of automated abuse detection \cite{Wulczyn2017, Founta2018}. The \textbf{wiki} dataset can be downloaded from \url{https://github.com/ewulczyn/wiki-detox} and is licensed under Apache License, Version 2.0. The \textbf{tweets} dataset can be downloaded with tweet ids from \url{https://github.com/ENCASEH2020/hatespeech-twitter}. These datasets cover two different domains: Wikipedia and Twitter. Each dataset is cleaned by removing extra white space, dropping duplicates and converting usernames, URLs and emoji to special tokens.

We fine-tune distil-roBERTa using the \texttt{transformers} integration with the \texttt{small-text} python package \cite{Wolf2019, Schroder2021smalltext}. distil-roBERTa has six layers, 768 hidden units, and 82M parameters. We encode input texts using the distil-roBERTa tokenizer, with added special tokens for usernames, URLs and emoji. All models were trained for 3 epochs with early stopping based on the cross-validation loss, a learning rate of $2e-5$ and a weighted Adam optimizer. All other hyperparameters are set to their \texttt{small-text} defaults. In each active learning iteration, we use 10\% of each labeled batch for validation. 
As a baseline to transformers-based AL, we use a support vector machine with no pre-training which we implement with \texttt{sklearn}. To encode a vector representation of input texts, we use a TF-IDF transformation fitted to the training dataset. 

All experiments were run on the JADE-2 cluster using one NVIDIA Tesla V100 GPU per experiment. For transformer-models, it took on average 1.5 hours to run each experiment. For SVM, it took less than a minute to run each experiment and these can be easily be run on a CPU. We repeat each experiment three times using three seeds to initialize a pseudo-random number generator.
\begin{table}
\centering
\footnotesize
\caption{The effect of varied keyword density thresholds on F1, false positive rate (FPR) and false negative rate (FNR).}
\label{tab:app_kws}
\begin{tabular}{c|ccc} 
\toprule
\multicolumn{1}{c}{\textbf{K}} & \textbf{F1}                                & \textbf{FPR}                              & \textbf{FNR}                                \\ 
\hline
\multicolumn{1}{c}{}           & \multicolumn{3}{c}{wiki}                                                                                                             \\ 
\hline
1.0\%                          & {\cellcolor[rgb]{0.58,0.831,0.706}}76.0\%  & {\cellcolor[rgb]{0.729,0.89,0.812}}2.7\%  & {\cellcolor[rgb]{0.561,0.82,0.694}}52.8\%   \\
5.0\%                          & {\cellcolor[rgb]{0.69,0.875,0.784}}69.0\%  & {\cellcolor[rgb]{0.412,0.761,0.588}}0.5\% & {\cellcolor[rgb]{0.741,0.894,0.82}}71.8\%   \\
10.0\%                         & {\cellcolor[rgb]{0.341,0.733,0.541}}91.0\% & {\cellcolor[rgb]{0.357,0.737,0.553}}0.1\% & {\cellcolor[rgb]{0.89,0.957,0.925}}87.4\%   \\
25.0\%                         & 49.0\%                                     & {\cellcolor[rgb]{0.341,0.733,0.541}}0.0\% & {\cellcolor[rgb]{0.996,0.996,0.996}}98.4\%  \\ 
\hline
                               & \multicolumn{3}{c}{Tweets}                                                                                                           \\ 
\hline
1.0\%                          & {\cellcolor[rgb]{0.439,0.773,0.608}}85.0\% & 4.5\%                                     & {\cellcolor[rgb]{0.341,0.733,0.541}}29.6\%  \\
5.0\%                          & {\cellcolor[rgb]{0.514,0.804,0.663}}80.0\% & {\cellcolor[rgb]{0.761,0.902,0.831}}2.9\% & {\cellcolor[rgb]{0.463,0.78,0.627}}42.7\%   \\
10.0\%                         & {\cellcolor[rgb]{0.471,0.784,0.631}}83.0\% & {\cellcolor[rgb]{0.475,0.784,0.631}}0.9\% & {\cellcolor[rgb]{0.788,0.914,0.851}}76.4\%  \\
25.0\%                         & {\cellcolor[rgb]{0.596,0.835,0.718}}75.0\% & {\cellcolor[rgb]{0.361,0.741,0.553}}0.2\% & 98.5\%                                      \\
\bottomrule
\end{tabular}
\end{table}
\section{Sampling with Keywords}
\label{sec:app_keywords}
We use a heuristic to weakly label examples from the unlabeled pool to be selected for the initial seed. Keywords are a commonly-used approach \cite[e.g. see][]{EinDor2020} and searching for text matches is computationally efficient over a large pool of unlabeled examples. However, the keyword heuristic only approximates the true label and can introduce biases due to non-abusive use of offense and profanities. In our data, we rely on a keyword density measure ($K$) which equals the number of keyword matches over the total tokens in a text instance. We then experiment with varied thresholds of $K \in$ [1\%, 5\%, 10\%, 25\%] for a weak label of abusive text. A higher threshold reduces false positives but also decreases true positives. We find a threshold of 5\% best balances these directional effects. Making predictions using a keyword heuristic with a 5\% cut-off achieves an F1-score relative to the true labels of 69\% for wiki and 80\% for tweets. Using this threshold, examples are expected to be abusive if the percentage of keywords in total tokens exceeds $5\%$. We then sample equal numbers of expected abusive and non-abusive examples from the pool, reveal their true labels and initialize the classifier by training over this seed.

\section{Additional Experimental Analysis}
\label{sec:app_additional}
\begin{figure*}
        \centering
        \begin{subfigure}[b]{0.475\textwidth}
            \centering
            \includegraphics[width=1.1\textwidth]{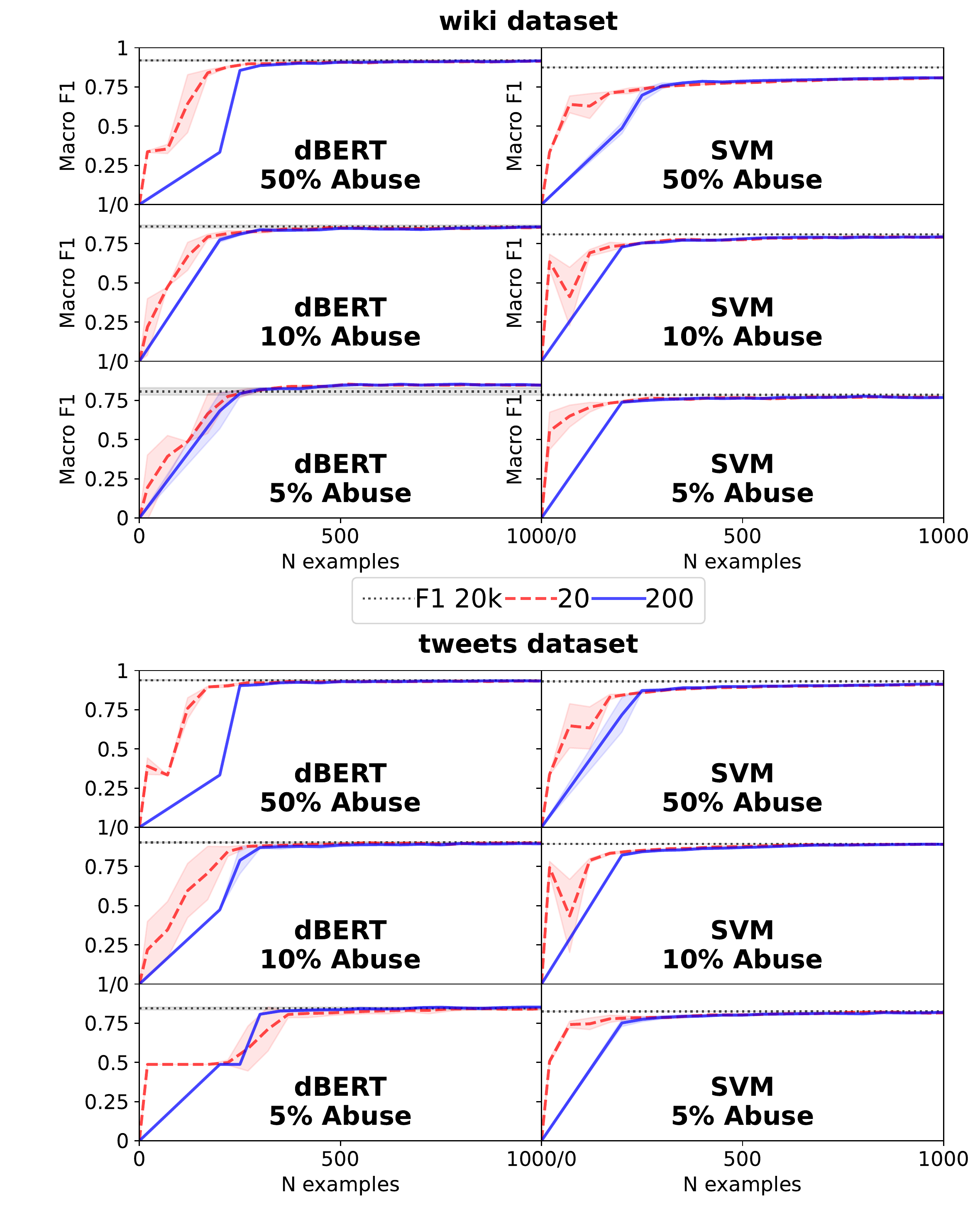}
            \caption{Seed Size}
            \label{fig:app_seed}
        \end{subfigure}
        \hfill
        \begin{subfigure}[b]{0.475\textwidth}  
            \centering 
            \includegraphics[width=1.1\textwidth]{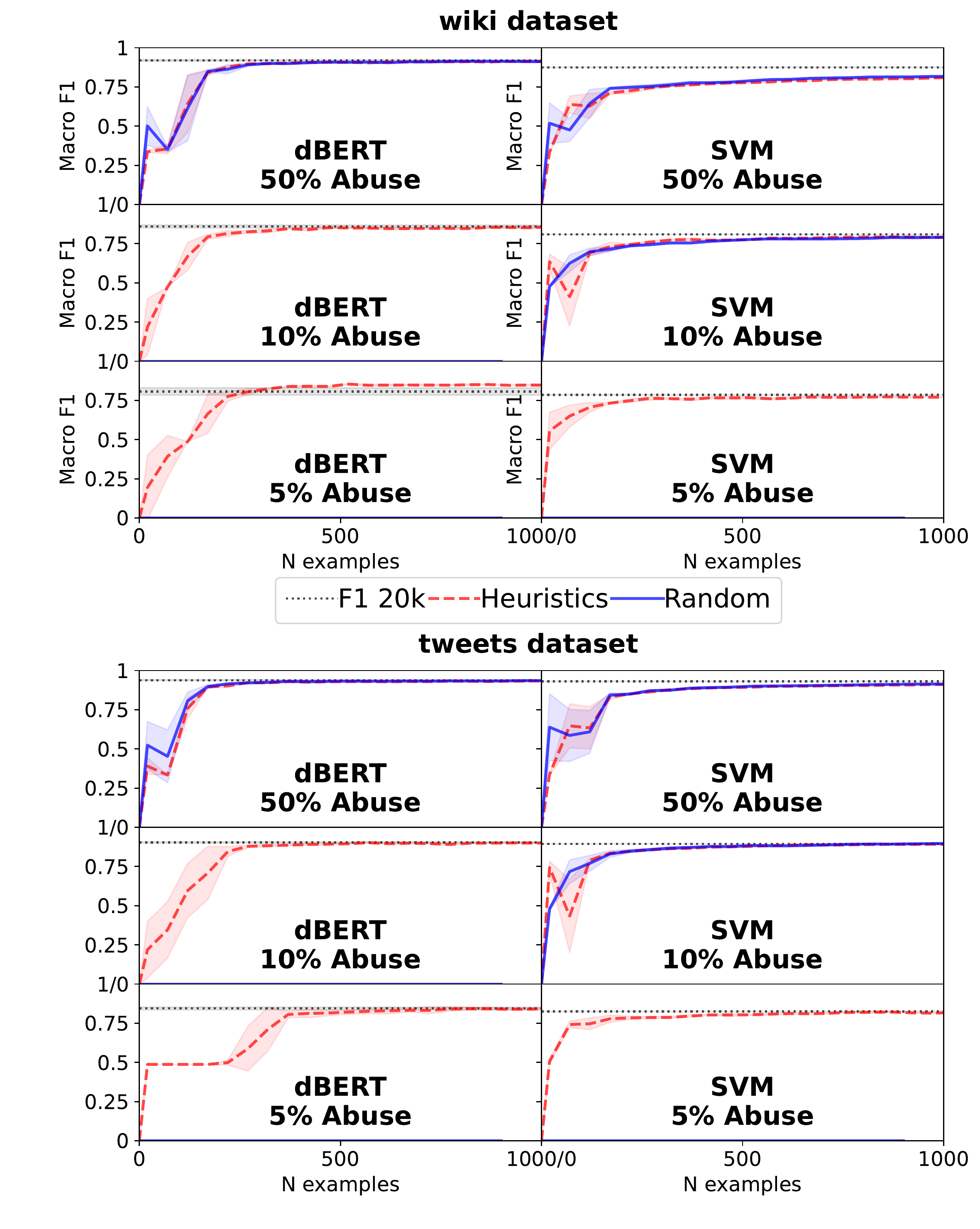}
            \caption{Cold Strategy}  
            \label{fig:app_cold}
        \end{subfigure}
        \vskip\baselineskip
        \begin{subfigure}[b]{0.475\textwidth}   
            \centering 
            \includegraphics[width=1.1\textwidth]{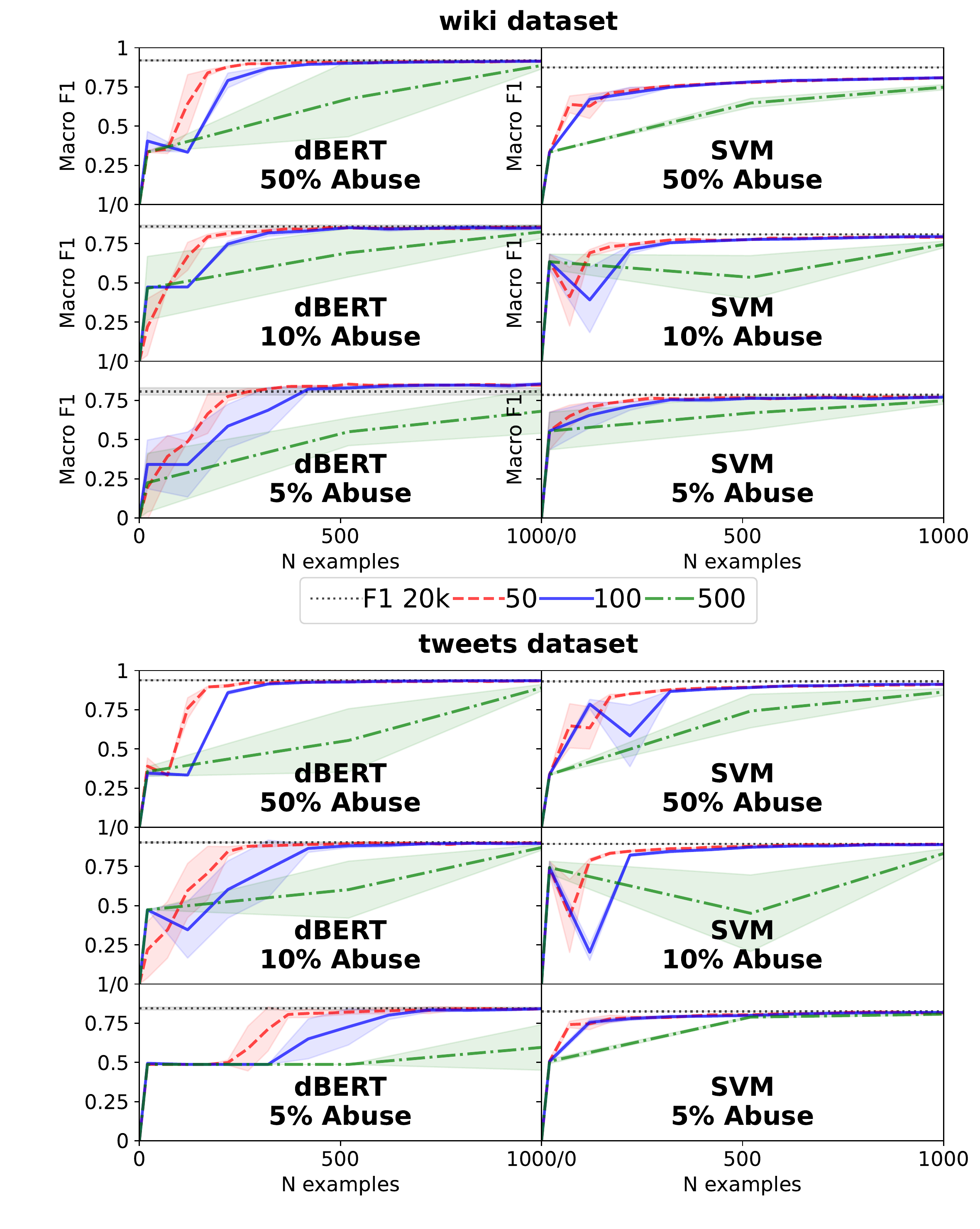}
            \caption{Batch Size}  
            \label{fig:app_batch}
        \end{subfigure}
        \hfill
        \begin{subfigure}[b]{0.475\textwidth}   
            \centering 
            \includegraphics[width=1.1\textwidth]{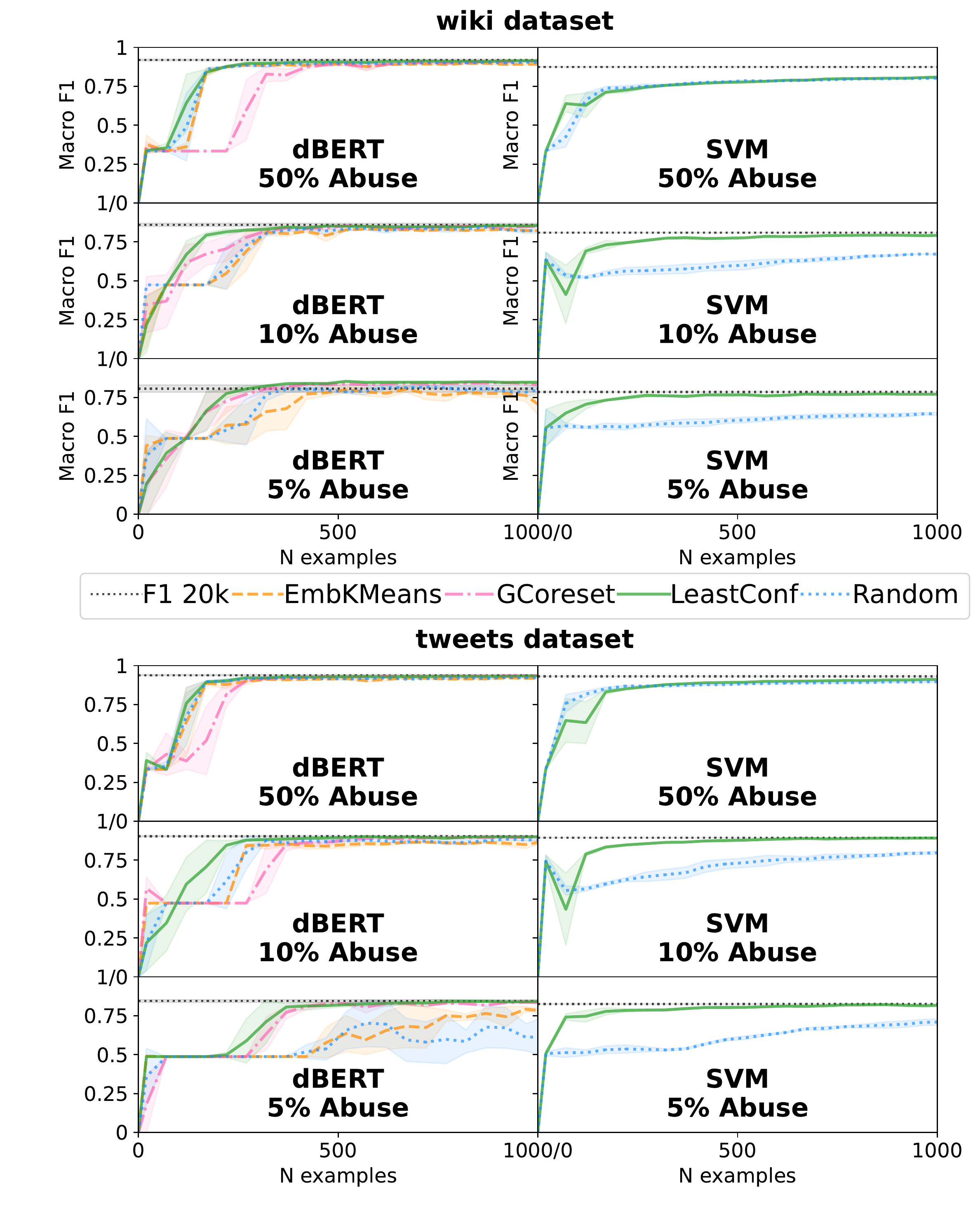}
            \caption{Query Strategy}    
            \label{fig:app_query}
        \end{subfigure}
        \caption{Learning curves per dataset-class imbalance pair showing the effect of isolated experimental variables on traditional (SVM) and transformers-based (dBERT) active learning.} 
        \label{fig:app_learning_curves}
    \end{figure*}

\begin{table}[ht!]
\centering
\caption{The best AL parameters and performance for each classifier (transformers vs SVM).}
\label{tab:best_table}
\footnotesize
\setlength{\tabcolsep}{1pt}
\resizebox{\columnwidth}{!}{
\begin{tabular}{cc|cccc|ccc} 
\toprule
\begin{tabular}[c]{@{}c@{}}\\\end{tabular} & \textbf{}           & \multicolumn{4}{c|}{\underline{\textbf{Best AL Combinations$^*$}}}              & \multicolumn{3}{c}{\underline{\textbf{Metrics}}}                                 \\
\textbf{Dataset}                           & \textbf{Classifier} & \textbf{Seed} & \textbf{Cold} & \textbf{Batch} & \textbf{Query} & \multicolumn{1}{c}{\textbf{$\mathbf{F1_{20k}}^\dagger$}} & \textbf{$\mathbf{F1_{AL}}$} & \textbf{$\mathbf{N_{90}}$}  \\ 
\hline
\multirow{2}{*}{\textbf{wiki50}}           & dBERT              & 20            & Random        & 50             & LC      & \textbf{0.920}                      & \textbf{0.922} & \textbf{170}  \\
                                           & SVM                 & 20            & Random        & 50             & LC      & 0.875                               & 0.838          & 1520          \\ 
\hline
\multirow{2}{*}{\textbf{wiki10}}           & dBERT              & 20            & Heuristic     & 50             & LC      & \textbf{0.859}                      & \textbf{0.866} & \textbf{170}  \\
                                           & SVM                 & 20            & Heuristic     & 50             & LC      & 0.809                               & 0.810          & 320           \\ 
\hline
\multirow{2}{*}{\textbf{wiki5}}            & dBERT              & 20            & Heuristic     & 50             & LC      & \textbf{0.807}                      & \textbf{0.855} & 220           \\
                                           & SVM                 & 20            & Heuristic     & 50             & LC      & 0.785                               & 0.780          & \textbf{170}  \\ 
\hline
\multirow{2}{*}{\textbf{tweets50}}         & dBERT              & 20            & Random        & 50             & LC      & \textbf{0.939}                      & \textbf{0.938} & \textbf{170}  \\
                                           & SVM                 & 20            & Random        & 50             & LC      & 0.931                               & 0.926          & 220           \\ 
\hline
\multirow{2}{*}{\textbf{tweets10}}         & dBERT              & 20            & Heuristic     & 50             & LC      & \textbf{0.904}                      & \textbf{0.902} & 220           \\
                                           & SVM                 & 20            & Random        & 50             & LC      & 0.893                               & 0.901          & \textbf{170}  \\ 
\hline
\multirow{2}{*}{\textbf{tweets5}}          & dBERT              & 200           & Heuristic     & 50             & LC      & \textbf{0.844}                      & \textbf{0.856} & 300           \\
                                           & SVM                 & 20            & Heuristic     & 50             & LC      & 0.825                               & 0.830          & \textbf{170}  \\
\bottomrule
\multicolumn{1}{l}{\textit{Notes:}} & \multicolumn{8}{l}{$^\dagger$ global metric from passive training over the full dataset} \\
\multicolumn{1}{l}{} & \multicolumn{8}{l}{$^*$ calculated by averaging the rank performance on $\mathbf{F1_{AL}}$, $\mathbf{N_{90}}$}
\end{tabular}
}
\end{table}
\cref{tab:best_table} shows the best parameters for each dataset and each classifier. In \cref{fig:app_learning_curves}, we present the learning curve and comparisons of each experimental variable for both datasets and classifiers. In each panel of \cref{fig:app_learning_curves}, we vary one parameter whilst holding all others fixed. This allows us to evaluate the impact of one variable, \textit{ceteris paribus}. Namely, the reference values are those reported in the main paper: seed size of 20 selected by heuristics-based sampling and a batch size of 50 queried by LeastConfidence strategy. 

\paragraph{Seed and Batch Size} We test two choices for seed size (20, 200), and three choices for batch size (50, 100, 500). We find AL is more efficient with smaller seeds and batch sizes. The F1 score achieved with a seed of 20 and four AL iterations of 50 ($|\mathcal{L}|=220$) exceeds that reached with a seed of 200 and 0 iterations ($|\mathcal{L}|=200$) by 55pp for \textbf{wiki50}, 4pp for \textbf{wiki10}, and 10pp for \textbf{wiki5}. Batch sizes of 100 and 500 are less efficient than 50, with 700--1,100 and 150--200 more examples needed for $\mathbf{N_{90}}$, respectively.

\paragraph{Seed Acquisition Strategy (Cold)} We evaluate two choices to select the examples for the seed.
(1) \textbf{Random:} Seed examples are randomly selected. Depending on the class distribution of the unlabeled pool (which, in real world settings, is unknown) only non-abusive content might be identified. For datasets expected to be approximately balanced, a randomly-selected seed has a high probability of including both class labels.
(2) \textbf{Heuristics:} Seed examples are selected using keywords ($n=652$), taken from the abusive language literature \cite{hateoffensive, hatelingo, peertopeerhate, Gabriel2018}. For \textbf{wiki50}, random- and heuristics-based initialization achieve equivalent $\mathbf{N_{90}}$. However, with a seed of 20, a third of randomly-initialized experiments fail on \textbf{wiki10} and all experiments fail for \textbf{wiki5}. This shows that when the data is imbalanced, a random seed is sub-optimal because both class labels are not observed.

\paragraph{Query Strategy}
In addition to LeastConfidence (LC), we evaluate two further strategies coupled with dBERT: 1) \textbf{GreedyCoreSet} is a data-based diversity strategy which selects items representative of the full set \cite{Sener2017} and 2) \textbf{EmbeddingKMeans} is a data-based diversity strategy which uses a dense embedding representation (such as BERT embeddings) to cluster and sample from the nearest neighbors of the $k$ centroids \cite{Yuan2020}. On our datasets, these two strategies are high performing in terms of the maximum F1 score they achieve over $2{,}000$ examples, but take longer to learn and are less efficient than LeastConfidence.

\section{Generalizability of Performance}
\label{sec:app_generalize}
\begin{figure}[!b]
    \centering
    \begin{subfigure}[b]{\columnwidth}
        \centering
        \includegraphics[width=\linewidth]{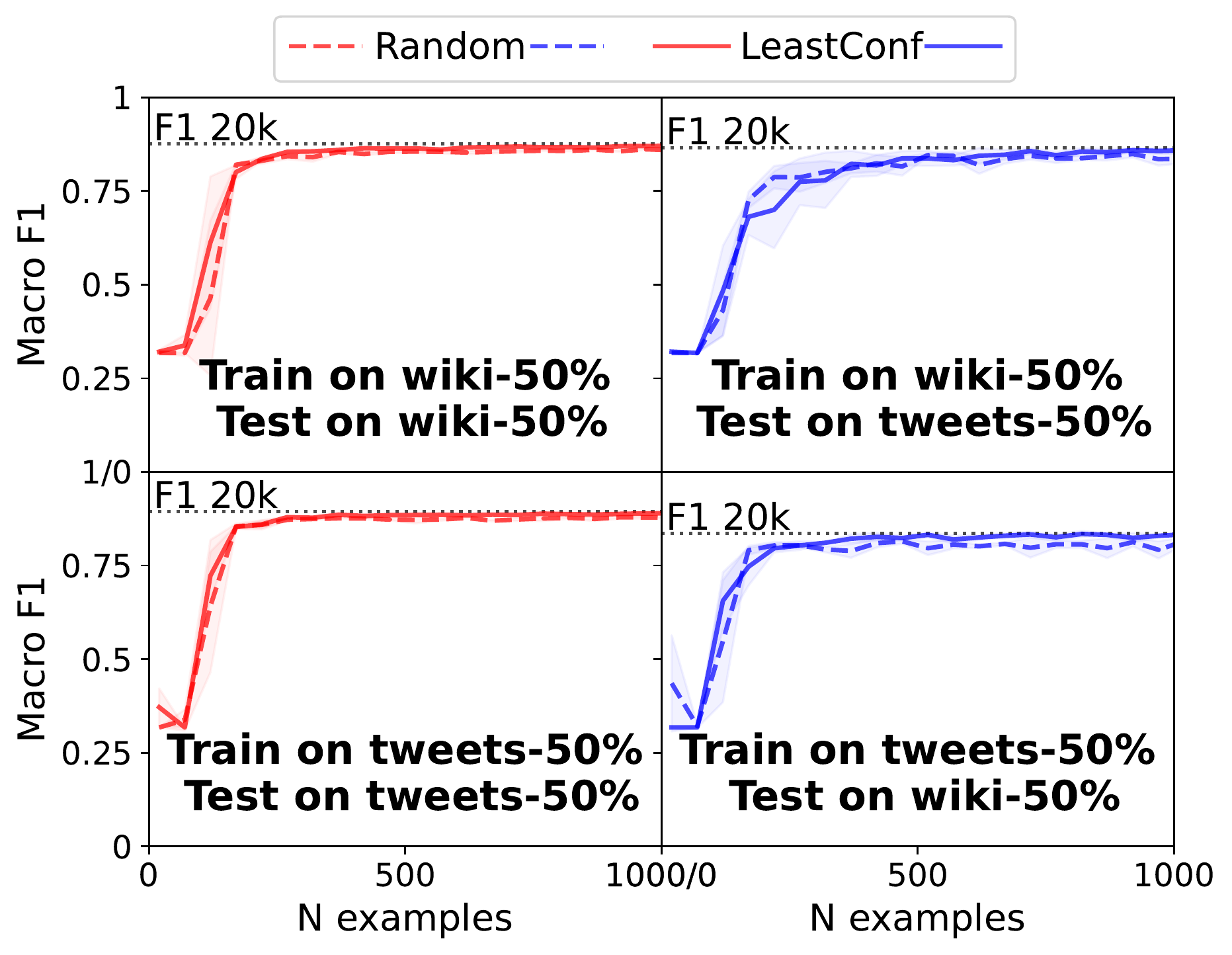}
        \caption{Models trained on 50\% abuse}
        \label{fig:50_cross}
    \end{subfigure}
    \vfill
    \begin{subfigure}[b]{\columnwidth}  
        \centering 
        \includegraphics[width=\linewidth]{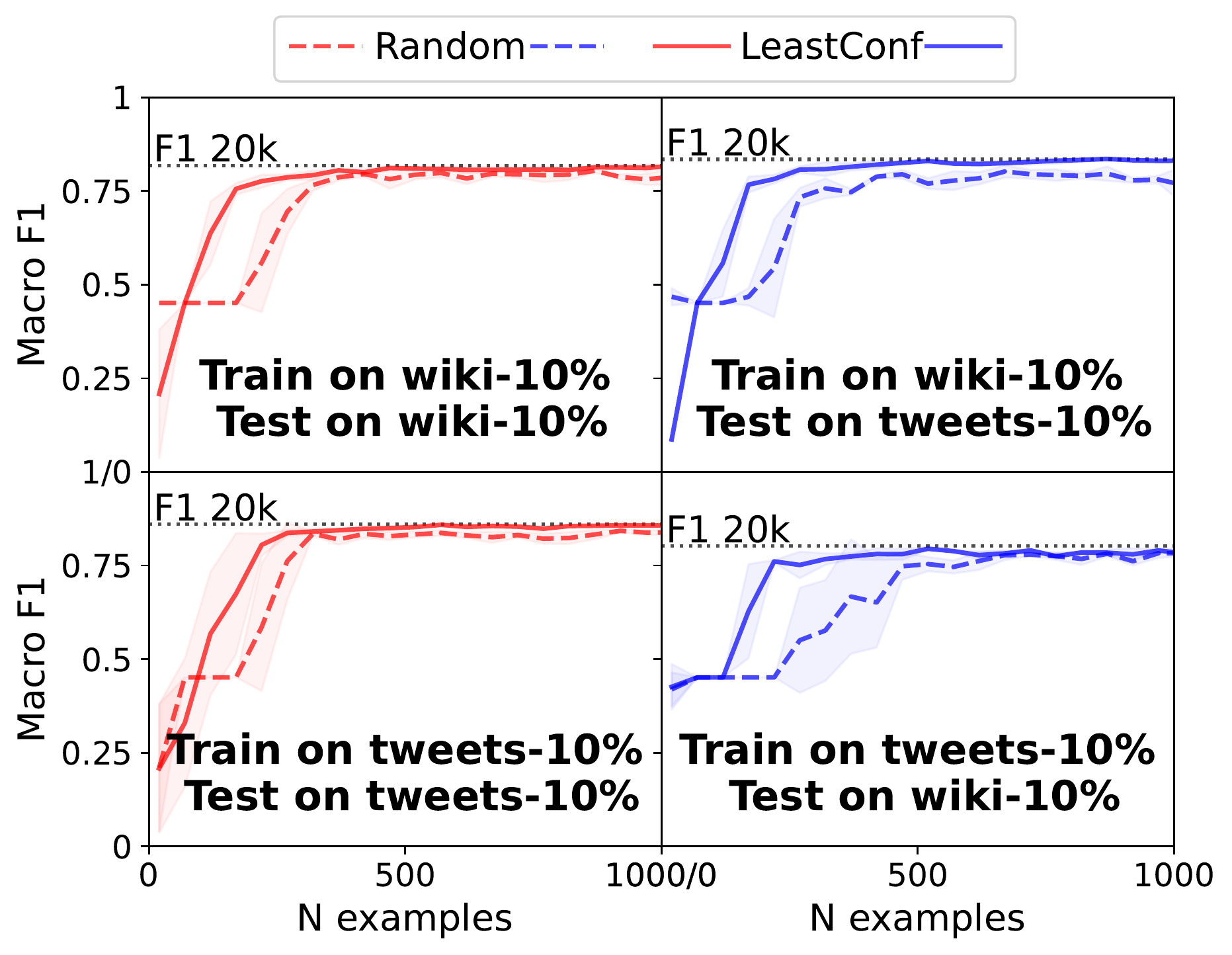}
        \caption{Models trained on 10\% abuse}  
        \label{fig:10_cross}
    \end{subfigure}
    \caption{Cross-dataset generalization (dBERT) for 50\% and 10\% abuse.} 
    \label{fig:app_generalize}
\end{figure}

In the main paper, we present the results of cross-dataset generalization with 5\% abuse. In \cref{fig:app_generalize}, we demonstrate the equivalent results for all class imbalances and both datasets. In general, tweets is harder to predict than wiki, so we see a larger change in performance when training on tweets and evaluating on wiki. For 50\% and 10\% abuse, performance is similar across test sets. For 5\% abuse, there is a larger difference especially for the random baseline. However, in all cases, the performance of the LeastConfidence strategy generalizes well to out-of-domain testing, at least for these two datasets which are similar in their proportion of explicit abuse \cite{wiegandDetectionAbusive2019}.

\end{document}